\documentclass[letterpaper, 10 pt, conference]{ieeeconf}

\IEEEoverridecommandlockouts

\usepackage{amsmath} 
\usepackage{amssymb}  
\usepackage[dvipsnames]{xcolor}

\usepackage[font=small, labelfont=bf]{caption}
\usepackage{subfig}
\usepackage{graphicx}
\usepackage{tikz}
\usetikzlibrary{calc,shapes,arrows,decorations.text,positioning,patterns,matrix,chains,decorations.pathreplacing}
\usepackage{hyperref}
\usepackage{diagbox}
\usepackage{siunitx}

\usepackage{multirow}
\usepackage{color}
\makeatletter
\newcommand{\thickhline}{%
    \noalign {\ifnum 0=`}\fi \hrule height 1pt
    \futurelet \reserved@a \@xhline
}
\newcolumntype{"}{@{\hskip\tabcolsep\vrule width 1pt\hskip\tabcolsep}}
\makeatother

\usepackage{tablefootnote}
\usepackage{multicol, blindtext}

\title
{
\LARGE \bf
Semantically Meaningful View Selection
}

\author
{
\\[-3.0ex]
Joris Gu\'erin$^{1}$, Olivier Gibaru$^{1}$, Eric Nyiri$^{1}$, St\'ephane Thiery$^{1}$ and Byron Boots$^{2}$%
\thanks{$^{1}$Laboratoire d'Ing\'enierie des Syst\`emes Physiques Et Num\'eriques, Arts et M\'etiers ParisTech, Lille, France. Emails: \{joris.guerin; olivier.gibaru; eric.nyiri; stephane.thiery\}@ensam.eu}%
\thanks{$^{2}$School of Interactive Computing, Georgia Institute of Technology, Atlanta, GA 30332. Email: bboots@cc.gatech.edu}%
\\[-3.0ex]
}

\begin{document}

\maketitle
\thispagestyle{empty}
\pagestyle{empty}

\begin{abstract}
An understanding of the nature of objects could help robots to solve both high-level abstract tasks and improve performance at lower-level concrete tasks. Although deep learning has facilitated progress in image understanding, a robot's performance in problems like object recognition often depends on the angle from which the object is observed. Traditionally, robot sorting tasks rely on a fixed top-down view of an object. By changing its viewing angle, a robot can select a more semantically informative view leading to better performance for object recognition. In this paper, we introduce the problem of semantic view selection, which seeks to find good camera poses to gain semantic knowledge about an observed object. We propose a conceptual formulation of the problem, together with a solvable relaxation based on clustering. We then present a new image dataset consisting of around 10k images representing various views of 144 objects under different poses. Finally we use this dataset to propose a first solution to the problem by training a neural network to predict a ``semantic score'' from a top view image and camera pose. The views predicted to have higher scores are then shown to provide better clustering results than fixed top-down views.
\end{abstract}

\section{INTRODUCTION}

Recent advances in machine learning have increased robot autonomy, allowing them to better understand their own state and the environment, and to perform more complex tasks. In particular, improving the semantic understanding of objects is an important research topic, which can aid in tasks such as manipulation. For example,  semantic information can directly help in solving tasks such as supervised \cite{rgbd_objectrec, fastrcnn_sorting} and unsupervised \cite{cnnClustering} sorting. It can also indirectly impact other important tasks, such as robotic grasping \cite{grasping_survey, deep_grasp}. The way people grasp objects depends on not only the form and shape of the object, but also on their semantic understanding of the object \cite{grasp_knowledge}. Knowledge of manipulated objects is especially important for human robot collaboration \cite{hrc_assembly}, where  robot behavior should be safe and adapt to human requirements~\cite{hrc_adapting}. For example, a robot should not hand a human a knife by the blade.

\input{fig_pb.tex}

Given recent advances in deep learning for both supervised \cite{xception, vgg} and unsupervised \cite{jule, survey_deepclu} image classification, vision-based methods are a natural choice for acquiring knowledge about manipulated objects. In contrast to most computer vision problems, robotic vision can leverage the robot's actuators  to change the  view under which an object is observed. This can have a huge impact on understanding what the object is. For example, in Figure~\ref{fig:dataset}, only the middle image enables the robot to understand that it is looking at a comb. The robot's ability to act has not been fully exploited in previous research. For example, prior work on robotic sorting of objects relies on a fixed, perpendicular top pose for the robot camera~\cite{cnnClustering, fastrcnn_sorting}. While there has been some previous work for best view selection \cite{bestviewselection}, this has focused on producing representative views of 3d mesh models. 
Although this is a promising approach, it is not applicable for many robotics tasks, especially when complete 3d models are not available for all of the manipulated objects. 

In this paper, we aim to find a method to optimize the poses of a robot with a hand-mounted camera, to maximize the semantic content of image and understand the nature of the objects being observed (Figure \ref{fig:dataset}). In Section \ref{sec:prob_desc}, we first propose a generic conceptual formulation of this problem, which we call the Semantic View Selection Problem (SVSP). We then relax the problem by reducing it to the optimization of a clustering-based objective. To solve this problem, we introduce a new image dataset containing 144 objects, from 29 categories, under different poses and observed under various views. 
Both the data collection process and the dataset content are described in Section \ref{sec:dataset}. A first approach using the clustering SVSP formulation on the new dataset is then detailed in Section \ref{sec:approach}. It consists in training a multi-input deep convolutional neural network to map a top view and proposed camera pose to a semantic score. Our experimental results, in Section \ref{sec:results}, demonstrate that the proposed network can predict camera poses which outperform fixed poses on unsupervised sorting tasks.
\section{THE SEMANTIC VIEW SELECTION PROBLEM} \label{sec:prob_desc}

In this section, we formally introduce the problem of selecting optimal views for semantic understanding and introduce notation used throughout the rest of the paper.

\subsection{Generic formulation: the semantic function}
\label{sec:generic}

Given an object $o$ in pose $p_o$, a view $v_{p_o, p_{cam}}$ is defined by the pair $(p_o, p_{cam})$, and represented by the image produced by the camera, where $p_{cam}$ is the pose of the camera. Thus, given $p_o$, there exists a direct mapping between the space of all possible views and the space of reachable camera poses. For each view, we define the conceptual \emph{semantic function} $S(.)$, representing the semantic information contained in a view. $S(v_{p_o, p_{cam}})$ is high if $v_{p_o, p_{cam}}$ is highly informative about the object being observed (second row of Figure \ref{fig:dataset}) and low if it's not (third row of Figure \ref{fig:dataset}). More concretely, semantic meaningfulness can be viewed as the information contained in the output of a high level feature extractor (e.g. last layer of a pretrained deep CNN). It can be used to infer the general category of the object represented in the image. Given $o$ in pose $p_o$, the Semantic View Selection Problem (SVSP) aims to find $p_{cam}^*$ such that the view $v_{p_o, p_{cam}^*}$ maximizes $S$.

\subsection{First relaxation: clusterability functions}
\label{sec:relax1}

The semantic function defines the general form of the SVSP, but, in practice, it cannot be evaluated. Therefore, we introduce a new family of \emph{clusterability functions} $\{S^{c, m}, c \in C, \, m \in M\}$, where $C$ represents the space of all possible image clustering pipelines and $M$ the space of all clustering evaluation metrics. In other words, an element of $C$ is a function mapping any set of images to a corresponding set of labels. Likewise, an elements of $M$ are functions that take two sets as inputs (predicted labels and ground truth labels) and output a real valued score, usually in $[0, 1]$. As explained later, we assume that if $c$ and $m$ are  a good image clustering routine and a good clustering metric respectively, then $S^{c, m}$ and $S$ are highly positively correlated. 



Let $O_{\infty}$ be the conceptual infinite set of all possible objects, $P^o_{\infty}$ be the space of all possible poses of object $o$, and $V^{p_o}_{\infty}$ be the space of all possible views of object $o$ in pose $p_o$. An image clustering problem with $N$ images is defined by $cp = \{v_{p_{o[i]}} ~ | ~ \forall i \in \{1,...N\}, o[i] \in O_{\infty}, p_{o[i]} \in P^{o[i]}_{\infty}, v_{p_{o[i]}} \in V^{p_{o[i]}}_{\infty}\}$. In other words, any set of images containing an underlying label can be viewed as a clustering problem, although these labels are not necessarily known. There are various ways to define the category of an object. In this paper, we use the most generic and simple possible label (e.g. spoon, mug, toothbrush, ...), without adding any specific description (e.g. silver spoon, blue mug, ..). Let $\mathcal{P}_{cp, c}$ be the cluster assignments (predictions) for $cp$, under clustering routine $c$ and let $\mathcal{L}_{cp}$ be the ground truth labels associated with $cp$. Then, for $m \in M$, we define $m_{cp, c} = m(\mathcal{P}_{cp, c}, \mathcal{L}_{cp})$. Finally, we define $\mathcal{CP}_{\infty}^{v}$ the infinite set of all possible clustering problems containing view $v$ (i.e. the set of all sets of images containing $v$). Then, $S^{c, m}$ is defined by
\begin{equation}
    S^{c, m}(v) = \mathop{\mathbb{E}}_{cp \in \mathcal{CP}_{\infty}^{v}}[m_{cp, c}],
\end{equation}
which is the average score under metric $m$ of all possible clustering problems containing $v$. $S^{c, m}(v_{p_o, p_{cam}})$ is high if the view is good for clustering $o$ and low if not, where good means having a high score under $m$.

We assume that $c$ and $m$ are a good image clustering routine and a good clustering metric respectively, i.e. they have been shown to work well in practice. Then, the assumption of high positive correlation between $S^{c, m}$ and $S$ is based on the intuition that a semantically meaningful image should be properly clustered with similar objects by a good clustering pipeline. Indeed, the clustering pipeline used in our experiments consists of extracting features from a pretrained deep CNN and clustering the new set of features using a standard algorithm. This choice for $c$ is in line with the definition of semantic meaningfulness proposed in Section \ref{sec:generic}, as the final representation of a view, passed to the clustering algorithm, is a vector a features extracted from a pretrained CNN. Another motivation for choosing a clustering-based estimate for the semantic function is that supervised classification or object detection methods might not be adapted. Indeed, to compute the Monte-Carlo estimate of such function (see Section \ref{sec:finiteData}), the selected algorithm needs to be run many times on relatively small datasets. Doing this in a supervised way has high chances to result in overfitting, in which case all views would have high semantic scores.

\subsection{Second relaxation: clusterability on a finite dataset}
\label{sec:finiteData}

As it is not feasible to consider all possible objects, poses,
and views, we further relax the above definitions to consider
a finite dataset. Let $O_{N}$ be a finite set of objects containing $N$ elements. As in Section \ref{sec:relax1}, we define $P^o_{N_{o}}$ a set of $N_{o}$ poses of object $o$, and $V^{p_o}_{N_{p_o}}$ a set of $N_{p_o}$ views of object $o$ in pose $p_o$. In other words, the set $\mathcal{D} = \{v_{p_o} \, | \, o \in O_{N}, p_o \in P^o_{N_{o}}, v_{p_o} \in V^{p_o}_{N_{p_o}}\}$ is an image dataset containing $\sum_{o \in O_{N}} \sum_{p_o \in P^o_{N_{o}}} (N_{p_o})$ images. If the dataset is large and diverse enough, an estimate of $S^{c, m}(v)$ can be computed by
\begin{equation}
    S_{\mathcal{D}}^{c, m}(v) = \mathop{\mathbb{E}}_{cp \in \mathcal{CP}_{\mathcal{D}}^{v}}[m_{cp, c}],
\end{equation}
where $\mathcal{CP}_{\mathcal{D}}^{v}$ is defined like $\mathcal{CP}_{\infty}^{v}$ with $cp$s sampled from $\mathcal{D}$.

For large datasets, it might be computationally intractable to compute $S_{\mathcal{D}}^{c, m}(v)$ as the number of possible combinations of images grows exponentially with the number of views. Thus, we propose to compute the Monte-Carlo estimate 
\begin{equation}
    \hat{S}_{\mathcal{D}}^{c, m}(v) = \mathop{\mathbb{E}}_{cp \in \mathcal{CP}_{\mathcal{D}, MC}^{v}}[m_{cp, c}],
\end{equation}
where $\mathcal{CP}_{\mathcal{D},MC}^{v}$ is a subset of $N_{MC}$ elements of $\mathcal{CP}_{\mathcal{D}}^{v}$, and $N_{MC}$ is a large natural integer ($N_{MC} \geq 2 \times 10^5$ in our experiments).

\subsection{Partially-observable Semantic View Selection }
\label{sec:partiallyobs}

Given an object $o$ in pose $p_o$, the relaxed SVSP aims to find a camera pose $p_{cam}$ such that $\hat{S}_{\mathcal{D}}^{c, m}(v_{p_o, p_{cam}})$ is high. In a generic robotic pipeline, the exact pose of an object is unknown and needs to be estimated from partial observations. Let $\omega_{p_o}$ be the observation from which we want to compute $\hat{p}_o$, the estimate of $p_o$. For example, $\omega_{p_o}$ can be a top-view image, taken from an initial predefined camera pose. Our approximation of the clusterability function score is then dependent on $\omega_{p_o}$ as a surrogate for the exact pose. More concretely, we want to optimize the parameters $\alpha$ of a function $f_{\alpha}: \{\omega_{p_o}, p_{cam}\} \rightarrow s \in D_m$, where $D_m$ is the output domain of the metric $m$, such that $s$ is an estimate of $\hat{S}_{\mathcal{D}}^{c, m}(v_{p_o, p_{cam}})$. A typical practical choice for $f_{\alpha}$ would be a convolutional neural network (CNN), where $\alpha$ represents its trainable parameters.
\section{DATASET CONSTRUCTION}\label{sec:dataset}

To tackle the proposed relaxed SVSP, we have built an image dataset\footnote{The dataset can be downloaded at \url{https://github.com/jorisguerin/SemanticViewSelection_dataset}.} representing various everyday objects under different poses, and observed under multiple views with a camera mounted on the end-effector of a UR10 robot manipulator (see Figure \ref{fig:dataset}). The dataset statistics can be found in Table \ref{tab:dataset}.

\begin{table}[!ht]
\caption{Statistics of our multi-objects/multi-pose/multi-view image dataset.} 
\label{tab:dataset}
\centering

    \begin{tabular}{c|c|c|c}
        
        \multirow{2}{*}{\# Classes} & \# Object/class & \# Poses/object & \# Images/pose \tabularnewline
        & (\textcolor{Blue}{\textit{total}}) & (\textcolor{Blue}{\textit{total}}) & (\textcolor{Blue}{\textit{total}}) \tabularnewline \hline
        
        29 & 4-6 (\textcolor{Blue}{\textit{144}}) & 3 (\textcolor{Blue}{\textit{432}}) & 17-22 (\textcolor{Blue}{\textit{9112}}) \tabularnewline
        
    \end{tabular}
   \vspace{-10pt}
\end{table}

\subsection{Estimating object location and size}

The dataset was collected using an Asus Xtion RGBD sensor, hand-mounted on a UR10 robot manipulator. For a given object $o$ in a given pose, we gather images corresponding to several views, with $o$ centered in the image. The first step is to estimate the location of the Geometrical Center of the object ($GC_{o}$). To do so, we place the robot in an initial pose such that the camera can see the entire workspace in which objects can be placed. We store a background image of this pose, corresponding to what the camera sees when there is no object. Then, using RGB background subtraction, the $xy$-contour of the object is obtained (the $z$ axis is vertical). From this contour, we estimate the $x$ and $y$ components of $GC_{o}$, the width and the length of $o$. Finally, we compare the minimum values of the point cloud inside and outside the $xy$-contour to estimate both the $z$ component of $GC_{o}$ and the height.

\subsection{Parameterization of camera poses} 

To parameterize camera poses, we define a reference frame at $GC_{o}$. We then compute $d = \sqrt{length^2 + width^2 + height^2}$, the diagonal of the object's bounding box, and define the radius $R$ such that $d$ takes $70\%$ of the smallest dimension of the image if the optical center of the camera ($OC_{cam}$) is at a distance $R$ of $GC_{o}$ and $z_{cam}$ is pointing towards $GC_{o}$. The camera poses are sampled on the half-sphere of radius $R$, centered at $GC_{o}$, such that $z_{o}$ is positive. For each position of $OC_{cam}$ on the sphere, the camera is positioned such that $z_{cam}$ is pointing towards $GC_{o}$, $x_{cam}$ is in the $xy_{o}$ plane and $y_{cam}$ is pointing ``upwards''.

\begin{figure}
    \centering
    \includegraphics[width = 0.34\textwidth, height = 0.306\textwidth]{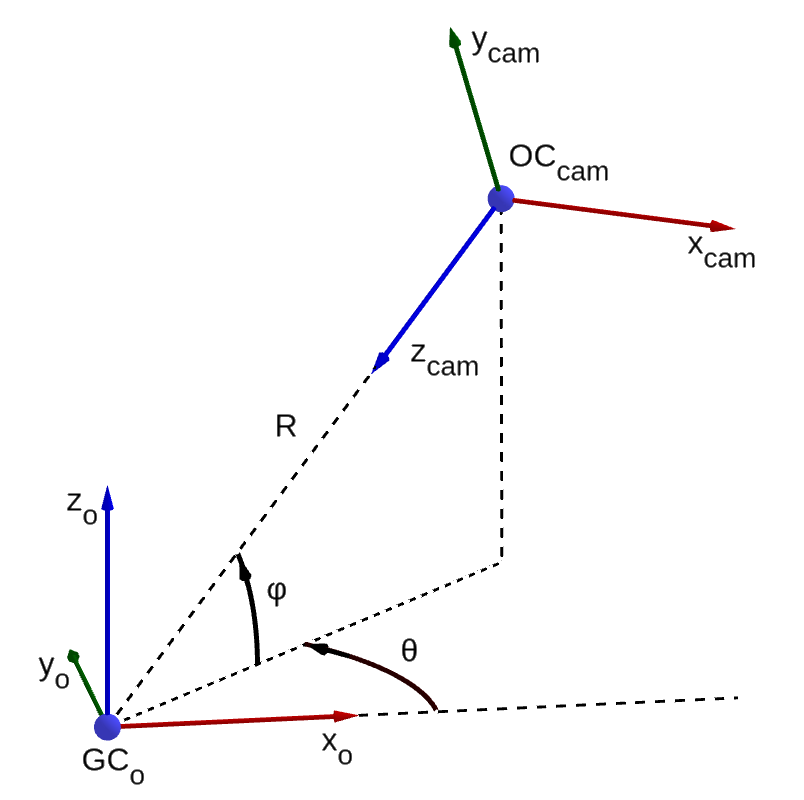}
    
    \caption{Definition of the parameters used to sample camera poses ($R$, $\theta$ and $\varphi$).}
    
    \vspace{-10pt}
    \label{fig:parameterization}
\end{figure}

\subsection{View sampling and data collection}

On the sphere, the location of $OC_{cam}$ is localized by two angles, $\theta$ and $\varphi$, which are defined as in Figure \ref{fig:parameterization}. Hence, a camera pose is simply represented by a ($\theta$, $\varphi$) pair. In our implementation, $\theta$ is sampled every $\ang{45}$ between $\ang{0}$ and $\ang{315}$, $\varphi$ is sampled every $\ang{15}$ between $\ang{45}$ and $\ang{75}$. The views for $\theta = \ang{270}$ correspond to configurations where the camera is oriented towards the robot base. They were not collected in the dataset to avoid seeing the robot on the images. The other missing values come from unreachability of the camera poses with the robot manipulator, which occurs when the RRT connect \cite{rrtconnect} planner fails to generate a valid plan. Furthermore, while it could be interesting to sample angles lower than $\varphi = \ang{45}$, these configurations are often unreachable because the robot would collide the table. A subset of the views gathered for one object in a particular pose can be seen in Figure \ref{fig:sampleim}.

\input{data_samples.tex}
\section{PROPOSED APPROACH}\label{sec:approach}

\subsection{Clustering pipeline and metric}

Given a view $v$, the clusterability function $\hat{S}_{\mathcal{D}}^{c, m}(v)$ used to represent the semantic function, are defined by both a good clustering pipeline $c$ and a clustering evaluation metric $m$. 

In this work, we use the image clustering pipeline proposed in \cite{cnnClustering}, which consists of getting a new representation of each image from the last layer of a deep CNN feature extractor $fe$, pretrained on ImageNet, and then clustering the new set of features using a standard clustering algorithm $c'$. Although some variants of this algorithm are tested in Section \ref{sec:results}, the standard pipeline in this paper uses Xception \cite{xception} to extract features and agglomerative clustering to cluster the deep features set. We use the implementation and weights of Xception proposed by the Keras library.

The clustering metric chosen to represent the clusterability is the Fowlkes-Mallows (FM) index \cite{fm}, defined by
\begin{equation}
    FMI_{cp, c} = \frac{TP}{\sqrt{(TP + FP) (TP + FN)}},
\end{equation}
where $TP$, $FP$ and $FN$ respectively represent the number of true positive, false positive and false negative pairs after clustering $cp$ using $c$. The FM index ranges between $0$ and $1$. We choose this index because it can be converted straightforwardly to a local form
\begin{equation}
    FMI^i_{cp,c} = \frac{TP_i}{\sqrt{(TP_i + FP_i) (TP_i + FN_i)}}, 
\end{equation}
where $FMI^i$ represents the individual FM score of image $v_i \in cp$, $TP_i$, $FP_i$ and $FN_i$ respectively represent the number of true positive, false positive and true negative pairs containing $v_i$. This individual form is used in the next section to reduce the sample complexity for computing $\{\hat{S}_{\mathcal{D}}^{c, m}(v), v \in \mathcal{D}\}$. The clustering pipeline and metric being chosen, we drop the $c$ and $m$ superscipts from  now on.

\subsection{Training Set}

We start by isolating 5 categories  at random from the dataset (comb, hammer, knife, toothbrush, wrench). This way, objects that were neither used for fitting the clusterability indices nor for training the neural network can be used to validate the network outputs. In this section, $\mathcal{D}$ refers to all the views composing the $24$ remaining categories. From $\mathcal{D}$, a clustering problem is created by sampling randomly the number of categories, the selected categories, the number of objects per category, the selected objects, the pose for each object, and one view for each pose. As every estimated quantity is computed on the training set from now on, we drop the subscript $\mathcal{D}$. To build the training set for the SVSP, we start by generating a large set of $N_{cp}$ random clustering problems $\mathcal{CP}_{MC} = \{cp_i, i \in N_{cp}\}$. Then, for each $(cp, v)$ pair, we define the following intermediate score
\begin{equation}
\label{eq:scoreindiv}
    \Tilde{s}_{cp}^v =
    \begin{cases}
      FMI^v_{cp}, & \text{if}\ v \in cp \\
      0, & \text{otherwise}.
    \end{cases}
\end{equation}
The \emph{individual semantic view score} of view $v$ is then defined by $\hat{s}^v = \sum\limits_{cp \in \mathcal{CP}_{MC}} \Tilde{s}_{cp}^v / N_{cp}^v$, where $N_{cp}^v$ is the number of elements in $\mathcal{CP}_{MC}$ containing $v$. Likewise, we can define a \emph{global semantic view score} for $v$, $\hat{S}^v$ by replacing $FMI^v_{cp}$ by $FMI_{cp}$ in (\ref{eq:scoreindiv}).

We build $\mathcal{CP}_{MC}$ such that $\min(\{N^v_{cp}, v \in \mathcal{D}\})$ is at least $2 \times 10^5$. This requires to solve $N_{cp} \approx 3 \times 10^7$ clustering problems. Because of the high computational expenses of this process, we cannot get a much higher number of samples for the Monte Carlo estimate. Hence, $\hat{s}^v$ seems more appropriate than $\hat{S}^v$ to estimate the semantic content of $v$ because it evaluates the individual contribution of each view to the global clustering results.

Hence, our training set for the next section is composed of $\{\{v_{top}^v, \varphi^v, \theta^v\}, \hat{s}^v\}$ input/output pairs. $v_{top}^v$ represents the top view image associated with $v$, $\varphi^v$ and $\theta^v$ are the angles parameterizing $v$. We note that for each pose, the scores among all the views are scaled to the $[0,1]$ interval to help training, as some objects are harder to cluster than others. In such case the best views of a difficult object might have lower $\hat{s}$ values than the worst views of an easy object. This makes the view selection problem harder because predicting the intrinsic ``clusterability'' of an object from a poor view can be very challenging.

\subsection{Learning to predict semantic scores}

After computing semantic view scores for each view in our training set, we aim to solve the SVSP introduced in Section \ref{sec:partiallyobs}. To do so, we train a neural network to predict $\hat{s}^v$ from a triplet $(v_{top}, \theta^v, \varphi^v)$.

To solve this problem, we use a multi-input neural network architecture. The top image is first passed through a convolutional block (VGG \cite{vgg} architecture with 19 layers initialized with weights pretrained on ImageNet). Then, the output of the convolution block are fed into a first multi-layer perceptron (MLP). The outputs of this first MLP are concatenated with the angular inputs and fed into a second MLP, which outputs the semantic estimate. The architecture is summarized in Figure \ref{fig:network}, where BN denotes a batch normalisation layer, Drop($x$) a dropout layer with $x\%$ drops and FC($k$) a fully connected layer with $k$ neurons. ReLu and Sigmoid are standard activation layers. This architecture for Semantic View selection is referred to as SV-net.

\input{network.tex}

We used the Adam optimizer, with an initial learning rate of $10^{-3}$, to train the neural network. The network architecture was cross-validated by randomly removing two categories from the training set. At test time, all the dropout rates are set to $0$.

We have several motivations for solving the SVSP by predicting a score from proposed camera poses instead of regressing directly on these poses. First, in robotics, it may be impossible to plan a trajectory to some views. Therefore, it seems more relevant to only consider reachable views. Second, the SVSP may have many possible solutions, in which case there is no unique mapping between a top view and a good camera pose. The robot can choose a good, reachable view by considering the predicted scores. Finally, training a network by regressing directly to angles from the top view  would likely cause it to converge to the average of all good poses, which itself might not be a good pose.
\section{EXPERIMENTS}\label{sec:results}

\subsection{Baseline for comparison}

Given an object clustering problem, we define a \emph{view selector} as any process used to select the view of an object. To evaluate the quality of a view selector on a clustering problem, we compare its results under a certain $(c, m)$ pair against two baseline view selectors. The first method is usually implemented for autonomous robot sorting; it consists of observing the object from the top view, and is denoted ``TOP'' in our experiments. Another baseline view selector chooses a view uniformly at random among the possible views (denoted ``RAND''). 

We consider a view selector successful if it outperforms these two baselines. 
Computing the best fixed view for our problem would be too computationally expensive, so, instead, we compare against one arbitrary fixed view (TOP) and also against randomly selected views to ensure that the selected fixed view was not particularly weak.

We first evaluate the relevance of the proposed semantic scores in Section \ref{sec:eval_score}, and then evaluate the results obtained by the view selection network in Section \ref{sec:eval_net}.

\subsection{Evaluation of the individual semantic view score.}\label{sec:eval_score}

The individual semantic view scores are fit using a particular $(c, m)$ pair. Therefor, to evaluate it, we must test it on additional clustering pipelines and  metrics. We vary pipelines by changing both the deep feature extractor and the clustering algorithm. The three pipelines tested are denoted XCE\_AGG, VGG\_AGG and XCE\_KM, where XCE stands for Xception, VGG for VGG19, AGG for agglomerative clustering and KM for KMeans. As for the clustering metrics, we use the Fowlkes-Mallows index (FM), normalized mutual information (NMI) and cluster purity (PUR), which are three commonly used metrics to evaluate clustering algorithms when the ground truth is known.

We compare two view selectors against RAND and TOP. The first one, denoted OPT$_{ind}$, consists of choosing the view with the highest individual semantic view score ($\hat{s}^v$). The second one, denoted OPT$_{glob}$, chooses the view with the highest global semantic view score ($\hat{S}^v$). The results are reported in Table \ref{tab:index}. All scores are averaged over $10^4$ clustering problems.

\begin{table}[!ht]
\caption{Semantic view scores validation. Comparison of clustering results among different view selectors on the training set. \textit{for each $(c, m)$ pair, the best view selector is in bold}.} 
\label{tab:index}
\centering
    \begin{tabular}{cc|ccc}

    & & FM & NMI & PUR\tabularnewline \hline
    \multirow{4}{*}{XCE\_AGG} & TOP & 0.48 & 0.78 & 0.73 \tabularnewline
    & RAND & 0.50 & 0.78 & 0.74 \tabularnewline
    & OPT$_{glob}$ & 0.85 & 0.94 & 0.93 \tabularnewline
    & OPT$_{ind}$ & \textbf{0.87} & \textbf{0.95} & \textbf{0.94} \tabularnewline
    \cline{1-5}
    \multirow{4}{*}{XCE\_KM} & TOP & 0.44 & 0.75 & 0.71 \tabularnewline
    & RAND & 0.46 & 0.76 & 0.72 \tabularnewline
    & OPT$_{glob}$ & 0.81 & \textbf{0.93} & 0.91 \tabularnewline
    & OPT$_{ind}$ & \textbf{0.83} & \textbf{0.93} & \textbf{0.92} \tabularnewline
    \cline{1-5}
    \multirow{4}{*}{VGG\_AGG} & TOP & 0.39 & 0.72 & 0.67 \tabularnewline
    & RAND & 0.38 & 0.72 & 0.66 \tabularnewline
    & OPT$_{glob}$ & 0.49 & 0.78 & 0.73 \tabularnewline
    & OPT$_{ind}$ & \textbf{0.52} & \textbf{0.79} & \textbf{0.74} \tabularnewline
    \end{tabular}
   \vspace{-7pt}
\end{table}

The first thing to note is that both semantic estimators, although fit with $c = $ XCE\_AGG and $m$ = FM, seem to pick views which are much better than TOP and RAND. This is not surprising as these results were computed on the dataset used to compute the estimators. However, it strengthens the belief that $\hat{s}^v$ and $\hat{S}^v$ are good semantic function estimators as they generalize to other feature extractors, clustering algorithms, and metrics. Surprisingly, we also note that $\hat{S}^v$ and $\hat{s}^v$ performances are very similar. This might mean that the number of samples in the MC computation is sufficient for $\hat{S}^v$ to be a good estimator of $S^v$. However, in our experiments, we also acknowledge that $\hat{S}^v$ values are much closer to each other for the different views, which reveals that information about individual data is lost when considering the global estimator. The slightly better results of the $\hat{s}^v$ estimator, as well as its better separability, justifies its use for training SV-net.

Finally, we refer the reader back to Figure \ref{fig:sampleim}, where both high and low $\hat{s}^v$ value images have been outlined. This gives a qualitative validation of the index relevance for estimating the semantic function. Indeed, it is easier to tell that the robot is looking at sun glasses from the green-outlined images than from the red-outlined ones.

\subsection{Evaluation of the learned semantic view selector}\label{sec:eval_net}

To evaluate our semantic view selection network (SV-net), we adopt an approach similar to the one outlined in the previous section.   The SV-net view selector is compared against RAND and TOP under various configurations on the test set, which was not included for $\hat{s}^v$ computation, or for training SV-net. Results are averaged over $10^4$ clustering problems randomly sampled from the test set and are reported in Table \ref{tab:network}. We note that SV-net was able to predict views which are better than TOP and RAND, which is a remarkable result, as these kinds of objects where never encountered by the network before. SV-net is able to extract sufficient information from a single top view image to predict if a camera pose will provide good high-level features about the object. As a qualitative validation, four samples of predicted images are displayed in Figure \ref{fig:predex}. We also highlight that the absolute values of the clustering scores cannot be compared across tables. There is no guarantee that views exist that are able to reach similar clustering accuracy when different classes of objects are considered.

\begin{table}[!ht]
\caption{SV-net validation. Comparison of clustering results between different view selectors on the test set. \textit{for each $(c, m)$ pair, the best view selector is in bold}.} 
\label{tab:network}
\centering
    \begin{tabular}{cc|ccc}

    & & FM & NMI & PUR\tabularnewline \hline
    \multirow{3}{*}{XCE\_AGG} & TOP & 0.44 & 0.51 & 0.70 \tabularnewline
    & RAND & 0.48 & 0.56 & 0.74 \tabularnewline
    & SV-net & \textbf{0.55} & \textbf{0.63} & \textbf{0.78} \tabularnewline
    \cline{1-5}
    \multirow{3}{*}{XCE\_KM} & TOP & 0.44 & 0.51 & 0.70 \tabularnewline
    & RAND & 0.48 & 0.55 & 0.73 \tabularnewline
    & SV-net & \textbf{0.55} & \textbf{0.62} & \textbf{0.78} \tabularnewline
    \cline{1-5}
    \multirow{3}{*}{VGG\_AGG} & TOP & 0.46 & 0.53 & 0.71 \tabularnewline
    & RAND & 0.44 & 0.51 & 0.70 \tabularnewline
    & SV-net & \textbf{0.48} & \textbf{0.55} & \textbf{0.73} \tabularnewline
    \end{tabular}
   \vspace{-7pt}
\end{table}

\input{net_examples.tex}
\section{CONCLUSION}

In this paper, we introduced a new problem called semantic view selection. The SVSP consists of finding a good camera pose to improve semantic knowledge about an object from a partial observation of the object. We created an image dataset and proposed an approach based on deep learning to solve a relaxed version of SVSP. 

By fitting an index based on averaged view clustering quality and training a neural network to predict this index from a top view image, we show that it is possible to infer which view results in good semantic features. This has many practical applications including autonomous robot sorting, which is generally solved from top view images only. Indeed, one can use the SV-net to enhance any sorting robot with the ability to select better views to reduce sorting errors.

In future work, we plan to extend the problem to multi-view selection. Indeed, multi-view learning is particularly well suited for this problem as multiple viewpoints are often necessary to understand the nature of objects. Implementing such a multi-view approach in the formalism of the ``next best view selection'' literature \cite{nextBestView} would also allow us to detect when the top view is already good, thus avoiding unnecessary computation in these cases. We also plan to combine the SVS framework with image segmentation and robot grasping to implement a real-world unsupervised robot sorting.

\section*{\normalsize ACKNOWLEDGMENTS}
\footnotesize
This work was carried out under a Fulbright Haut-de-France Scholarship as well as a scholarship from the Fondation Arts et M\'etiers (convention No. 8130-01). This work is also supported by the European Union's 2020 research and innovation program under grant agreement No.688807, project ColRobot (collaborative robotics for assembly and kitting in smart manufacturing).


\bibliographystyle{IEEEtran.bst} 
\bibliography{biblio}

\end{document}